%% file: emnlp2021.tex
\pdfoutput=1

\documentclass[11pt]{article}

\usepackage{authblk}
\usepackage{xurl}

\usepackage[]{emnlp2021}

\usepackage{times}
\usepackage{latexsym}

\usepackage[T1]{fontenc}

\usepackage[utf8]{inputenc}

\usepackage{microtype}

\usepackage[export]{adjustbox}
\usepackage{dblfloatfix}
\usepackage{float}

\usepackage{graphicx, multirow}
\usepackage{todonotes}
\usepackage{subcaption}

\usepackage{tabularx}
\usepackage{array}
\usepackage{makecell}
\usepackage{float}

\newcommand{\App}{i\textsc{FacetSum}}
\newcommand{\demoUrl}{https://biu-nlp.github.io/iFACETSUM/WebApp/client/}
\newcommand{\linkDemo}{\url{\demoUrl}}
\newcommand{\linkCode}{\url{https://github.com/BIU-NLP/iFACETSUM}}
\newcommand{\sect}{\S}
\newcommand\avi[1]{}
\newcommand\alon[1]{}
\newcommand\eran[1]{}
\newcommand\ori[1]{}
\newcommand\rpc[1]{}
\newcommand\mbc[1]{}

\title{\App: Coreference-based Interactive Faceted Summarization \\for Multi-Document Exploration}

\newcommand*\samethanks[1][\value{footnote}]{\footnotemark[#1]}

\author[1]{\bf Eran Hirsch}
\author[1,2\Thanks{~~Equal contribution.}]{\bf Alon Eirew}
\author[1\samethanks]{\bf Ori Shapira}
\author[1]{\bf Avi Caciularu}
\author[1]{\bf Arie Cattan}
\author[1]{\\ \bf Ori Ernst}
\author[3]{\bf Ramakanth Pasunuru}
\author[4]{\bf Hadar Ronen}
\author[3]{\bf Mohit Bansal}
\author[1]{\bf Ido Dagan}
{
\makeatletter
\renewcommand\AB@affilsepx{~~~ \protect\Affilfont} \makeatother
\affil[1]{Bar-Ilan University}
\affil[2]{Intel Labs, Israel}
\affil[3]{UNC Chapel Hill}
\affil[4]{Peres Academic Center}
}
\affil[  ]{} 
\affil[  ]{\tt \{hirsch.eran, obspp18\}@gmail.com}
\affil[  ]{\tt alon.eirew@intel.com dagan@cs.biu.ac.il}

\begin{document}
\maketitle

\input{figures/figSystemSnapshot}

\input{sections/abstract}

\input{sections/intro}

\input{sections/interface}

\input{sections/backend}

\input{sections/usr_study}

\input{sections/related_work}

\input{sections/conclusion}

\input{sections/ethical}

\input{sections/acknowledgments}

\bibliography{custom}
\bibliographystyle{acl_natbib}


\appendix
\input{sections/appendix}

\end{document}

%% file: figures/figSystemSnapshot.tex
\begin{figure*}[t!]
    \centering
    \resizebox{\linewidth}{!}{
        \includegraphics{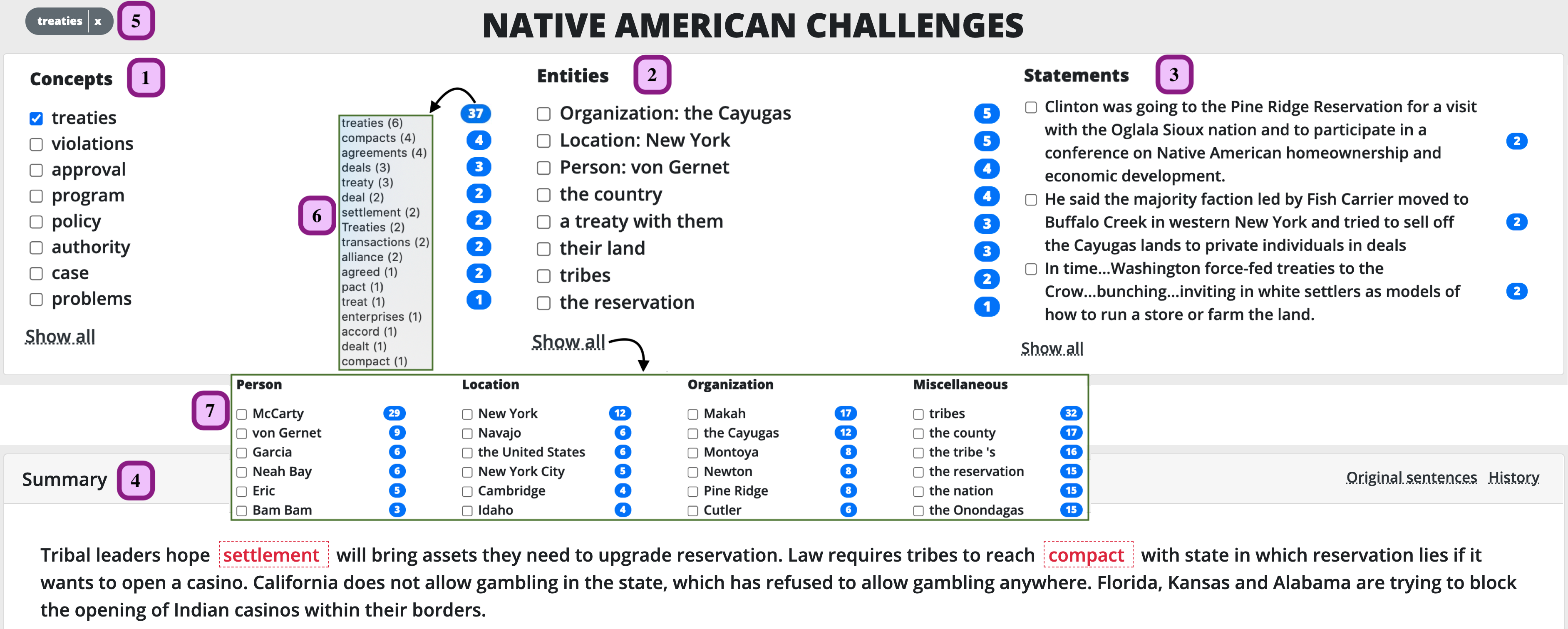}
    }
    
    \caption{Our \App{} web application over a set of 25 documents about ``Native American Challenges''. The user gets an overview of the topic as \textit{Concepts} [1], \textit{Entities} [2] and \textit{Statements} [3] facets. The facets are updated in response to the user's choice of the facet-value ``treaties'' [5]. An abstractive summary is generated for the set of sentences corresponding to the ``treaties'' semantic cluster [4].
    The mentions of a facet-value appear when hovering over its frequency [6]. Clicking "Show all" opens a pop-up with more facet-values. The \textit{Entities} pop-up is categorized into further facets of \textit{Person}, \textit{Location}, \textit{Organization} and \textit{Miscellaneous} [7].}

    \label{fig_systemSnapshot}
\end{figure*}

%% file: sections/abstract.tex
\begin{abstract}

We introduce \App{},\footnote{Demo at \linkDemo, and code at \linkCode.} a web application for exploring topical document sets.
\App{} integrates interactive summarization together with faceted search, by providing a novel faceted navigation scheme that yields abstractive summaries for the user's selections.
This approach offers both a comprehensive overview as well as concise details regarding subtopics of choice.
Fine-grained facets are automatically produced based on cross-document coreference pipelines,
rendering generic concepts, entities and statements surfacing in the source texts.
We analyze the effectiveness of our application through small-scale user studies, which suggest the usefulness of our approach.


\end{abstract}

%% file: sections/intro.tex
\section{Introduction}
\label{sec_intro}

An information consumer aspiring to explore a new topic will often be faced with an extensive collection of texts from which to acquire knowledge. Confronted with these texts, the reader would have difficulty determining where to start reading and obtaining details about specific aspects of the topic.
Addressing this we present \App, illustrated in Figure \ref{fig_systemSnapshot}, an \textit{interactive faceted summarization} approach and system for navigating within a large input document-set on a topic. The system initially provides a full high-level overview of the topic at a glance in the form of facets. A user can then dive further into subtopics of interest and obtain concise facet-based summaries, capturing the valuable information of a subtopic.
\mbc{should this first para also mention `coref' somewhere?}

The challenge of knowledge navigation has been addressed with various solutions, mainly under the umbrella of \textit{exploratory search} \citep{marchionini2006exploratorySearch} tasks. For example, in Complex Interactive Question Answering (ciQA) \citep{kelly2007overviewTrec} and Conversational QA \citep{reddy2019coqa}, a user interacts with a QA system in order to meet an information need on the source text(s). Interactive information retrieval \citep{ingwersen1992iir} and conversational search \citep{Radlinski2017convSearch} refine document retrieval through different means of textual interaction.
Both tasks do not offer a preliminary outline of the source documents, and hence expect a user to formulate queries or questions without system guidance. Furthermore, short answers, such as those output in conversational QA, may be insufficient, while lists of relevant textual results, such as in conversational search, may be overwhelming and provoke an inefficient navigation process.

As a midpoint solution, interactive summarization provides an initial summary as an overview of the topic, and the ability to inquire, via suggested or free-text queries, for more information in the form of summary expansions \citep[e.g.][]{shapira2021intSumm, avinesh2018sherlock}.
Here still, the initial summary, along with the suggested queries, do not produce the full high-level picture, and therefore hints only partially at the possible subtopics that the user might want to explore.

\App~builds upon the interactive summarization scheme, extending it via the effective faceted search approach \cite{hearst2006facetedDefinition} (\sect\ref{sec_interface:overview_component}), coupled with facet-based abstractive summarization (\sect\ref{sec_backend_summarization}). The presented facet values provide a comprehensive overview of the input topic, while the abstractive summaries deliver concise fine-grained information on selected facet values (see Figure \ref{fig_systemSnapshot}). Furthermore, since facets are hierarchically updated in accordance to facet-value selections, navigating deeper into subtopics becomes seamless.
In terms of backend implementation, facets are automatically derived over the input document set in a novel manner, based on cross-document coreference resolution \citep{cattan2021crossdocument} and proposition alignment \citep{ernst2020superpal}, yielding clusters of facet-value mentions (\sect\ref{sec_backend}).
Accordingly, summaries are generated based on the sentences that contain mentions of \textit{all} selected facet-values.
\mbc{should we explicitly mention the multi-facet `intersection' aspect too?}

We conduct usability studies on our system, and demonstrate its utility for easy navigation in topical document sets, while enabling deep diving into desired knowledge without losing the context of the exploration process.

We next describe \App's interface in \sect \ref{sec_interface} and its backend implementation in \sect \ref{sec_backend}. This is followed by the description and results of our usability investigations in \sect \ref{sec_experiments}, an overview of related work in \sect \ref{sec_relatedwork}, and finally conclusions and suggestions for future work in \sect \ref{sec_conclusion}.

%% file: sections/interface.tex
\section{\App~Interface}
\label{sec_interface}

\App~is a web application for exploring a document-set on a topic, shown in Figure \ref{fig_systemSnapshot}. It generally consists of the \textit{faceted navigation} component (top of figure, described in \sect \ref{sec_interface:overview_component})
, and the facet-based \textit{summary} component (bottom of figure, \sect\ref{sec_interface:summary_component}). The former rests upon a faceted-navigation panel that provides orientation on the source topic, while the latter supplies the user with key information about selected facet-values. This flow facilitates guided exploration, over the full scope of the topical information and within subtopics of interest.

\subsection{Faceted Navigation}
\label{sec_interface:overview_component}

Faceted search is a technique used to provide more effective information-seeking support \cite{tunkelang2009facetedsearch}, by allowing users to narrow down results based on rich attributes. A \textit{facet} describes an attribute type,
and \textit{facet-terms} or \textit{facet-values} represent attribute values. \App{'s} facets are formed using techniques that identify
recurring mentions
of sub-sentential units in texts, as explained further in \sect\ref{sec_backend_coref}.

The faceted navigation component is laid-out to the user in the form of three general facets (Figure \ref{fig_systemSnapshot}, [1], [2] and [3]): (1) \textit{Generic Concepts} facet,
e.g., ``poverty'' and ``treaties''.
(2) \textit{Entities} facet, 
containing values such as 
e.g., ``Clinton'' as a person or ``Nebraska'' as a location. (3) \textit{Statements} facet, which lists specific statements mentioned several times, such as ``Nebraska does not allow casino gambling''.

In our data scheme, each facet-value encapsulates a cluster of mentions that semantically refer to a common concept, entity or statement, and, as such, may be lexically diverse (e.g., the ``case'' concept associates with mentions of ``lawsuit'', ``fight'', ``battle'', ``debate'').
A \textit{facet-value sentence-set} is defined as the set of sentences pertaining to all of a facet-value's mentions.
The \textit{facet-value label} is the facet-value name presented to the user, and is chosen to be the most frequent lexical type in the mention cluster corresponding to that facet-value.

The values under each facet are ordered by their frequencies (number of mentions) in the source document set, as an indication for level of salience. A facet-value is shown with its corresponding frequency, and its various mention forms are revealed by hovering over the frequency meter (e.g., depicted in [6], the cluster ``treaties'' includes mentions of ``agreements'', ``deals'', etc.).
Only a few of the top facet-values are shown under each facet, while clicking \textit{Show all} expands the facet in full, in a pop-up. The pop-up for the \textit{Entities} facet partitions the facet-values to particular sub-facets: \textit{Person}, \textit{Location}, \textit{Organization} and \textit{Miscellaneous} ([7]).

By clicking a facet-value, the system generates a summary of its sentence-set. Additionally, the facets update to include only values appearing in that sentence-set. The updated facet view thus gives an overview which is fine-grained for the selected subtopic, while iteratively selecting additional facet-values supports diving deeper into it. When additional facets are gradually selected, a summary 
is generated over the intersection of the sentence-sets of all selected facets.
Any of the selected facet-values can be canceled out, whereby the system updates accordingly.

\subsection{Facet-based Summarization}
\label{sec_interface:summary_component}

Upon a change in selection of facet-values, the system provides the user with targeted information via an abstractive summary of the selections' sentence-set ([4]).
As more facet-values are selected, the generated summary is based on the intersection of the sentence sets of all selected facets, becoming more specific.
The user can further view the complete set of source sentences used to generate the summary, and those sentences' full documents (Figures \ref{fig_orig_sentences} and  \ref{fig_orig_document} in Appendix).
Additionally, clicking ``History'' shows all previously generated summaries (Figure \ref{fig_history} in Appendix).

%% file: sections/backend.tex
\section{Backend Algorithms}
\label{sec_backend}

\input{figures/figBackend}

As portrayed in \sect\ref{sec_interface}, \App{} supports two central features: presenting a faceted navigation panel and generating a summary around selected facet-values.
We next describe how facet-values are generated using CD coreference resolution (\sect \ref{sec_backend_coref}), and how we apply abstractive summarization, based on a facet-value selection (\sect\ref{sec_backend_summarization}).
Figure \ref{fig_backend} illustrates the entire process.

\subsection{Coreference-based Facet Formation}
\label{sec_backend_coref}
As described in \sect \ref{sec_interface:overview_component}, there are three main facets.
\textit{Concepts} and \textit{Entities} are extracted using cross-document (CD) coreference resolution pipelines, while \textit{Statements} via a proposition alignment pipeline, described next.\footnote{Facet extraction runs in a pre-processing step, since it is not fast enough for real-time latency (see Appendix \ref{sec_appendix_backend}).}

\paragraph{Concepts.}
We found that identifying and grouping together significant co-occurring events within the source document collection helps to expose and emphasize the notable concepts in the topic. To that end, we employ CD \textbf{event coreference resolution} which detects these concepts.

CD coreference resolution \cite{lee-etal-2012-joint} clusters text mentions that refer to the same event or entity across multiple documents.
Presently, the Cross-Document Language Model (CDLM) \cite{avicdlm2021} is the state-of-the-art for CD coreference resolution. This model is pretrained on multiple related documents via cross-document masking, encouraging the model to learn cross-document and long-range relationships.
Specifically, we employ the CDLM version fine-tuned for coreference on the ECB+ corpus \citep{cybulska-vossen-2014-using}. This model does not include a mention detection component, but rather expects relevant mentions to be marked within the input texts. We therefore leverage the mention detection ability of the 
model by \citet{cattan2021crossdocument}.

Once we have obtained the coreference clusters from CDLM, events whose mentions are predominantly verbs are filtered out,\footnote{\label{footnote_spacy} Using spaCy \citep{spacy2020}.} since those usually present specific actions that tend to be less informative compared to nominal types that refer to  more generic events (e.g., ``said'', ``found'' ``increase'' compared to ``unemployment'', ``poverty'', ``crash'').

CD event coreference resolution separates specific event instances, hence differentiating between clusters of similar event types with different arguments (e.g., ``unemployment'' in Navajo vs. ``unemployment'' in Cayuga).
Since generic event types, like ``unemployment'', are more suitable as facet-values, clusters with the same label (most frequent mention) are merged. Each such merged clusters then constitutes a single facet-value, to be presented to the user as described in \sect\ref{sec_interface:overview_component}.\footnote{We observed that the CD event coreference model has a tendency to wrongly collapse events of the same type, effectively aiding our concept formation.}

\paragraph{Entities.}
The \textit{Entities} facet-values help the user focus on entities such as people (e.g., "Clinton"), locations (e.g., "New York"), organizations (e.g., "FBI") and others (e.g., "the casino").
We created a separate pipeline for CD \textbf{entity coreference resolution}, since we observed subpar performance when applying the above CD coreference pipeline for entity coreference.\footnote{This is in line with previous work \cite{cattan2021crossdocument} which points out that the ECB+ dataset only considers entities that are arguments of event mentions, which is non-exhaustive.}

Unlike event coreference, mostly studied in the CD setting, entity coreference has recently seen impressive progress in the \textit{within}-document (WD) setting \cite{wu-etal-2020-corefqa, joshi-etal-2020-spanbert}. Hence, we leverage WD entity coreference in our entity recognition pipeline, which comprises three main steps. (1) We use SpanBERT\footnote{Using AllenNLP \cite{Gardner2017AllenNLP}.} \cite{joshi-etal-2020-spanbert}, a state-of-the-art transformer-based LM
for WD entity coreference resolution, to detect and cluster coreferring entity mentions within each separate document. (2) The entity mentions detected in the first step are marked as input for a CD entity coreference reolution model. To overcome ECB+ entity scarcity referred earlier, we use an alternative model that is trained on the WEC-Eng dataset \cite{eirew-etal-2021-wec}.\footnote{Fine-tuning CDLM on WEC-Eng is computationally infeasible, and therefore we use the model by \citet{eirew-etal-2021-wec}.} 
(3) Finally, we apply agglomerative clustering to combine the coreference clusters from steps 1 and 2 (WD and CD), and produce the overall entity coreference clusters (details in Appendix \ref{sec_appendix_backend}).

Once all entity coreference clusters are extracted, we bin them into more specific categories (``Person'', ``Location'' and ``Organization''), as portrayed in \sect\ref{sec_interface:overview_component}, by invoking a Named Entity Recognition (NER) model.\textsuperscript{\ref{footnote_spacy}} A facet-value cluster is tagged with the majority NER label of the mentions in the cluster, among Person, Organization and Location. If no NER label is assigned to a cluster, it is tagged as ``Miscellaneous'' (more details in Appendix \ref{sec_appendix_backend}).

\paragraph{Statements.}
Key statements benefit a user by presenting information about specific facts.
To generate these statements, we group together coreferring propositions (rather than words) that describe the same fact within the source documents, as seen in \sect\ref{sec_interface:overview_component}.

Following \citet{ernst2020superpal}, our pipeline consists of three steps. (1) Proposition candidates are extracted with OpenIE \citep{stanovsky-etal-2018-supervised}.
(2) Pairs of propositions expressing the same statement are matched using the SuperPAL model \citep{ernst2020superpal}, considering proposition pairs whose alignment score is above 0.5 as matched.
(3) A propositions graph is created by connecting pairs of nodes that represent similar propositions, and proposition clusters are matched for the connected components in the graph (more details in Appendix \ref{sec_appendix_backend}).

\subsection{Abstractive Facet Summarization}
\label{sec_backend_summarization}

In the standard summarization setting, a system receives a single or multiple documents as input, as well as a query in the query-focused task. In our case, the input is a set of sentences that have one or more selected facet-values in common, effectively providing a \textit{multi-facet} summary.
Given the set of sentences that correspond to the facet-value selection(s), these sentences are concatenated, ordered by their position in their source document (more details in Appendix \ref{sec_appendix_backend}). This text is then given as input to BART \cite{lewis2020bart}, a denoising sequence-to-sequence model fine-tuned on the single-document abstractive summarization task.\footnote{We use the huggingface model from \url{https://huggingface.co/facebook/bart-large-cnn}.}

\App{} presents abstractive rather than extractive summaries due to their enhanced readability, particularly when summarizing a set of related sentences.
This choice follows prior work, which showed that fusing sentences with shared points of coreference potentially facilitates coherence of abstractive summaries \citep{lebanoff2020corefSumm}. Indeed, in an internal manual assessment of 30 random individual summaries produced by \App, with 5 readability measures \citep{dang2006DUCoverview}, testers found overall that the summaries are highly readable.
To verify that factuality is not compromised, an additional inspection found that these summaries were also factually consistent to the input text, with 28 out of 30 sampled sentences marked as consistent.
See Appendix \ref{sec_appendix_summaryAssessment} for scores and more details on these assessments.

%% file: figures/figBackend.tex
\begin{figure*}[t!]
    \centering
    \resizebox{\textwidth}{!}{
    \includegraphics{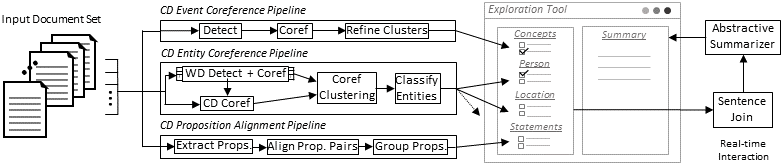}}
    \caption{The \App{} architecture. CD = cross-document, WD = within-document. \mbc{very blurry image? insert as cropped pdf}}
    \label{fig_backend}
\end{figure*}

%% file: sections/usr_study.tex
\section{System Experiments}
\label{sec_experiments}
\App{} aims to provide an effective means of information seeking in scenarios that require learning or investigating a new topic 
\citep{marchionini2006exploratorySearch}. 
To that end, we tested this goal through two small-scale experiments with human subjects, as a preliminary examination of the system. In the first experiment, we conducted a pilot usability study to inspect whether users felt they were able to satisfactorily complete an information seeking task using our system.
In the second, we examined whether \App{} is preferred over a standard document-search system to complete the exploration task.

\subsection{Usability Study}
\label{sec_experiments_userstudy}

\paragraph{Setup.}

The purpose of this experiment was to get general feedback, from human subjects, on the usability of the system, following established usability study methodologies \citep{nielsen1994usability}.
To simulate a realistic use case of topic exploration,
we instructed participants to use the system in order to prepare a draft review, given an informational goal, that a reporter could then use to write a report on the topic.
We prepared guiding story-lines (Appendix \ref{sec_appendix_experiments}, Table \ref{table_experiments_tasks}), as informational goals, for two topics from the DUC 2006 MDS dataset \cite{nist2006DUC}. To analyze \App~in different
exploratory situations, one topic is broad with higher information variability across the articles (``Native American Challenges''), while the other is more focused on a specific event (``EgyptAir Crash'').

In this pilot usability study, six participants\footnote{The discount usability testing principle contends that six evaluators are sufficient for prototype evaluation \citep{Nielsen1993Usability}.} explored both topics in random order.
During system usage we observed the users' activity, via a ``think aloud'' technique \citep{van1994thinkaloud}, to obtain user remarks. After exploring a topic, a user rated, from 1 to 5, the usefulness of different aspects of each component in the interface. After both topics, a System Usability Scale (SUS) questionnaire \cite{brooke1995sus} was filled, to assess global usability of the system (overall score from 0 to 100).
Further details are available in Appendix \ref{sec_appendix_usability_study}.

\paragraph{Results.}

The average SUS score over the 6 participants is 82.9, where $80.3$ is considered ``excellent'' \citep{uiux2021susScale}.
From the average component ratings over the 12 sessions, 
users expressed their satisfaction with the facet view's and summaries' quality for the use of the tasks. The overall facets quality received a score of 4.3 (SD=0.7), summary coherence 4.7 (SD=0.5), summary informativeness 4.2 (SD=1.1), summary non-redundancy 3.8 (SD=1.0), and summary length 4.3 (SD=0.9).
General feedback and issues raised by participants are available in Appendix \ref{sec_appendix_user_study_issues}. Overall, participants were pleased with their experience and some voiced their desire to use the tool right away for current event issues, like COVID-19 vaccination.

As expected, users noticed a difference between the two topics, and mentioned that they preferred the \textit{Concepts} facet for ``Native American Challenges'', while preferring the \textit{Entities} facet for ``EgyptAir Crash''.
Users found the \textit{Statements} facet-values rather lengthy and less useful, and at times considered it a substitute for summarizing the topic. Future improvements of the system may include considering alternative uses of the aligned statements, like linking specific fact mentions across documents.




\subsection{Comparative Analysis}
\label{sec_experiments_timing}

To further investigate whether \App{} is an effective tool for exploring a new topic, we conducted a small-scale comparison with a search tool, which roughly simulates common means for learning about a new topic. We asked four new experimentees to carry out the exploration task described in \sect\ref{sec_experiments_userstudy}, once with our system on one topic, and once with the search tool on the other topic (in different orders). The search tool used was DocFetcher,\footnote{\url{http://docfetcher.sourceforge.net}} an open source desktop search application, which indexes the given files, enables searching documents with queries, and highlights query terms within retrieved documents.
The participants finished their assignment with \App{} slightly faster than with the search tool. More importantly, they conveyed their satisfaction of using \App{} as a tool for navigating through multiple texts, and learning about a new topic.
The participants filled a questionnaire, rating each question on a scale of 1 (DocFetcher is preferred) to 7 (\App{} is preferred). The questions included:
(1) Which system was \textit{easier to use} in order to get the desired result? (Avg=5.5, SD=1.73); (2) With which system was it easier for you to \textit{get an overview} of the topic? (Avg=5, SD=2.3); (3) With which system was it easier for you to get \textit{detailed information} about a subtopic of interest? (Avg=5.25, SD=0.9); (4) If you had to learn about or explore a new topic, \textit{which system would you choose}? (Avg=5.25, SD=0.95). 
Overall, participants favored \App{} in all questions, preferring it for future use (details in Appendix \ref{sec_appendix_comparison_experiment}).

%% file: sections/related_work.tex
\section{Related Work}
\label{sec_relatedwork}

Attaining information of interest from large document sets has been approached with different techniques. 
A vast amount of research has been conducted on \textit{multi-document summarization}, as a method for presenting the central aspects of a target set of texts \citep[e.g.][]{barzilay1999mdsAbs, haghighi2009mds, bing2015mdsAbs, yasunaga2017graphMds}, 
where \textit{query-focused summarization} \citep{dang2005duc05} biases the output summary around a given query \citep[e.g.][]{daume2006qfs, baumel2018qfs, xu2020qfs}.

Recognizing the need for dynamically acquiring a broader or deeper scope of the source texts, \textit{exploratory search} \citep{marchionini2006exploratorySearch, white2009exploratorySerach} was coined as an umbrella term for allowing more dynamic interactive exploration of information.
Adapting the summarization paradigm to the exploratory setting, \textit{interactive summarization} enables a user to refine or expand on a summary via different modes of interaction. For example, \citet{shapira2021intSumm}, \citet{avinesh2018sherlock} and \citet{baumel2014qcfs} provide a limited (or no) initial summary on the document set, and support iterative interaction, via queries or preference highlights, to update the summary.
However, the succinct initial summary, possibly accompanied by few suggested queries, do not display the \textit{full} scope of the source texts, which limits the user's perception of the many available sub-topics to learn more about.

On the other hand, other exploratory search approaches \textit{do} provide a more elaborate overview of the source data through sophisticated dashboards or facets of extracted information or metadata \citep[e.g.][]{OConnor2010tweetmotif, koren2008personalizedFaceted, hope2020scisight}. Indeed, faceted navigation \citep{hearst2006facetedDefinition, tunkelang2009facetedsearch} is an effective instrument for navigating within a large data source \citep{hearst2002search, ruotsalo2020facetedMethod}.
While most faceted search systems generate facets from semi- or fully-structured data, as prominently encountered in e-commerce websites and in research \citep{hearst2006facetedDesign, BenYitzhak2008faceted}, some works generate facet hierarchies from unstructured open-domain texts. For example, from product reviews, \citet{ly2011reviewFaceted} extract product aspects and present several summaries, each focused on a single aspect as a ``facet", in a form of single-level faceted search.
\citet{hope2020scisight} devise facet-values from scientific articles by eliciting unstructured textual information (topics, entities) from the articles and their structured metadata (e.g article authors).
Although these search tools offer a more comprehensive overview of the source data, they either present raw-text search results or do not allow thorough navigation.

\App{} fully integrates dynamic multi-level faceted navigation into interactive multi-document summarization. The facets serve as an efficient means of grasping the topic, and render an intuitive medium for navigating through the information. The abstractive summaries generated at real-time expose concise details for any combination of sub-topics of choice.
Furthermore, we innovatively employ coreference resolution and proposition alignment to generate fine-grained open-domain facets.

%% file: sections/conclusion.tex
\section{Conclusion and Future Work}
\label{sec_conclusion}

In this paper, we presented \App{}, a novel text exploration approach and tool over large document sets, which incorporates faceted search into interactive summarization.
Its faceted navigation design provides a user with an overview of the topic and the ability to gradually investigate subtopics of interest, communicating concise information via multi-facet abstractive summarization.
Fine-grained facet-values are generated from the source texts based on cross-document coreference pipelines.
Small-scale user studies suggest the utility of our approach for exploring a new topic from multiple documents.

Future work may speed up the coreference-based facet extraction pipeline, allowing for real-time processing of ad-hoc document sets, and may investigate further methods for facet generation.
Additional search techniques might be integrated into the exploration scheme, including free text searching as raised in the user study. It would also be appealing to try adapting the system to domains other than news, such as the medical or scientific domains, for which exploration tools would be very useful. Such adaptations would depend on the portability of the underlying technologies of cross-document coreference resolution and proposition alignment. Finally, future work may explore additional ways of leveraging the power of recent proposition alignment methods.

%% file: sections/ethical.tex
\section{Ethical Considerations}
\label{sec_appendix_ethical}

\paragraph{Usability Study.}
We conducted the usability study (\sect\ref{sec_experiments_userstudy}) over Zoom sessions (\url{https://zoom.us/}), and carried out the ``think aloud'' technique through screen sharing and a with an open camera. Participants volunteered to take part in the study, and took about 45 minutes of each of their time. An \textit{informed consent} form was signed by the participant before each study.

The comparative study included four NLP doctoral students from our lab who volunteered for the experiment. The summary readability and factual consistency assessments were done by two authors of this paper.

\paragraph{Computation.}
We ran the three pre-processing pipelines mentioned in \sect\ref{sec_backend_coref} on 2 to 4 GPUs, where each pipeline ran from a few minutes to 10 hours per topic (25 news articles). Six such topics were prepared for the demo applications. (more details in Appendix \ref{sec_appendix_backend}).

The summarization model runs in real-time (per user interaction) over a CPU in less than 3 seconds per summary. Summaries are cached to refrain from recomputing summaries for repeated queries.


\paragraph{Dataset.}
The DUC 2006 data was acquired according to the required NIST guidelines ({\url{duc.nist.gov}}).

\paragraph{Multilingualism.} All models used within the components of \App{} were trained on English data, thus making the system compatible for English only. Supporting other languages requires replacing the contained models to ones compliant to the desired languages.

%% file: sections/acknowledgments.tex
\section*{Acknowledgments}
This work was supported in part by the German Research Foundation through the German-Israeli Project Cooperation (DIP, grant DA 1600/1-1); by the Israel Science Foundation (grant 1951/17); by a grant from the Israel Ministry of Science and Technology; and by grants from Intel Labs. RP and MB were supported by NSF-CAREER Award 1846185 and a Microsoft PhD Fellowship.

%% file: sections/appendix.tex
\section{Implementation Details}
\label{sec_appendix_implementation}

\subsection{Interface}
The frontend uses the reactjs library ({\url{https://reactjs.org/}}) and the bootstrap library ({\url{https://getbootstrap.com/}}).

\subsection{Backend}
\label{sec_appendix_backend}
The backend service is written in python, using the tornado web server library ({\url{https://www.tornadoweb.org}}). The summarization model was downloaded from huggingface (\url{https://huggingface.co/facebook/bart-large-cnn}). The service is deployed on a Linux server with CPU only.


All coreference and proposition alignment models described in \sect\ref{sec_backend_coref} are previously trained models. Links to these trained models are available in the project's GitHub.

For creating the CD coreference clusters for events with the fine-tuned CDLM model, we used two 32GB V100-SMX2 GPUs, for about 6 hours per topic.
For creating the CD coreference clusters for entities, we used one 12GB TITAN Xp GPU, for about 5 minutes per topic.
For creating the proposition alignment clusters we used four GeForce GTX 1080 Ti GPUs, for about 10 hours per topic.

\paragraph{CD entity coreference merging step.}

As described in \sect\ref{sec_backend_coref}, our final CD entity coreference step merges WD and CD predictions.
The SpanBERT WD model outputs clusters of coreferring mentions, while the CD entity model \cite{eirew-etal-2021-wec} outputs a pairwise score for each pair of mentions. We therefore convert SpanBERT clusters to mention-pair scores, by scoring pairs that are clustered together as 1, and 0 otherwise. Then, following common practice \cite{kenyon-dean-etal-2018-resolving, barhom-etal-2019-revisiting, meged-etal-2020-paraphrasing, eirew-etal-2021-wec}, we apply agglomerative clustering over all mention-pairs (both WD and CD) and produce the final entity coreference clusters.
Since WD coreference quality is superior to that of CD coreference, the high \textit{WD} coreferring mention-pair scores of 1 causes the clustering algorithm to favor those pairs for overall coreference clusters.

\paragraph{Proposition-level similarity threshold.}

The proposition alignment model computes a pairwise similarity between pairs of propositions, and we only consider pairs with a score above 0.5 (as a standard binary classification heuristic).
We then create a similarity graph, where each proposition is a node, and paired propositions are linked with an edge. The final clusters are the connected components in the graph. For example, if for propositions $P_1$, $P_2$ and $P_3$, there exist pairs $(P_1, P_2)$ and $(P_1, P_3)$, then $P_1$, $P_2$ and $P_3$ will be clustered together.

\paragraph{Facet-value label.}
As mentioned in \sect\ref{sec_backend_coref}, each facet-value is linked to a coreference cluster (a set of mentions) and has a label which is displayed to the user. For concepts and entities, this label is the text of the cluster's most frequent mention.
For statements, there is no repetition of mention texts in the cluster. There, we use the text of the longest mention, under the assumption that it has more context for the user to understand the statement.

\paragraph{Entities sub-facet categorization.}
After computing the \textit{Entities} facet-values with entity coreference resolution, we categorize each facet-value to a specific entity type.
For this, we first calculated the named entity class, with NER, for each mention in the facet-value cluster. All tokens of a mention were to be classified with the same NER class in order for the mention to be considered classified.
Then, the class repeating the most times in a cluster was chosen as the class of the cluster. If all mentions of a cluster were not classified, we categorized the facet-value as \textit{Miscellaneous}.

We mapped spaCy's NER classes to names that we found are more friendly to non NLP-practitioners (e.g., ``GPE'' is named ``Location'').

\paragraph{Facet-value filters.}
After generating the potential facet-values (coreference clusters), we filter out:
\begin{itemize}
  \addtolength\itemsep{-2mm}
  \item Clusters with more than 50 mentions, under the assumption that they are too noisy for the user.
  \item Singleton clusters, i.e. a cluster with one mention or one linked sentence (coreferring in the same sentence), under the assumption that they are uninformative.
  \item Clusters whose label is 2 characters or less (e.g., "'s", "AP").
  \item Clusters whose label has a verb part-of-speech tag.
\end{itemize}

\paragraph{Summarization model input.}
As described in \sect\ref{sec_backend_summarization}, BART is used to summarize the set of input sentences relevant to the facet-value selections.
Since BART has an input-length limit of 1024 tokens, ordering the sentences based on their sentence index raises the likelihood that summaries will be based on sentences from multiple documents. The documents were ordered by their alphanumeric file system order based on their document ID.

\subsection{Data}
DUC 2006 MDS dataset is used for demonstrating the application, specifically with 6 topics: D0601, D0602, D0606, D0608, D0617, D0629.

\section{Experiment Details}
\label{sec_appendix_experiments}

We carried out a usability study and a system comparison experiment (\sect\ref{sec_experiments}), as well as a summary quality evaluation (\sect\ref{sec_backend_summarization}).

\subsection{Usability Study}
\label{sec_appendix_usability_study}

For the usability study, six participants were gathered based on prior acquaintance.
Each user had a 45 minutes Zoom session with an experienced experiment overseer. The participants first filled an experiment participation consent form. Before starting the actual experiment, the users were presented with another topic for experimenting with the system, followed by instructions of the experiment overseer, to reduce the learning curve of using the system for the first time.

\input{figures/tabExperimentsTasks}

Table \ref{table_experiments_tasks} shows the two tasks that each user received. Participants conducted the experiments on the two topics in different orders.

\paragraph{SUS questionnaire.}
The SUS questionnaire \citep{brooke1995sus} was filled once by each user after both topics, with the following 10 questions:
\begin{enumerate}
\addtolength\itemsep{-2mm}
    \item I think that I would like to use this system frequently.
    \item I found the system unnecessarily complex. 
    \item I thought the system was easy to use.
    \item I think that I would need the support of a technical person to be able to use this system.
    \item I found the various functions in this system were well integrated.
    \item I thought there was too much inconsistency in this system.
    \item I would imagine that most people would learn to use this system very quickly.
    \item I found the system very cumbersome to use.
    \item I felt very confident using the system.
    \item I needed to learn a lot of things before I could get going with this system.
\end{enumerate}

To calculate the SUS score, the following procedure is taken \citep{brooke1995sus}: First sum the score contributions from each item. Each item's score contribution will range from 0 to 4. For items 1,3,5,7,and 9 the score contribution is the scale position minus 1. For items 2,4,6,8 and 10, the contribution is 5 minus the scale position. Multiply the sum of the scores by 2.5 to obtain the overall value of SU. SUS scores have a range of 0 to 100.

The final scores of the six participants are: 

\input{figures/tabSUSIndividual}

\paragraph{Usefulness questionnaire.}
After exploring each topic, the participants filled a questionnaire as follows:
\begin{itemize}
\addtolength\itemsep{-3mm}
	\item For the requirements of the given task, how useful was the Facets component between 1 (not useful at all) and 5 (very useful)?
	\item Overall, the summaries output by the system were: between 1 (I disagree) and 5 (I agree)
	\begin{itemize}
	\addtolength\itemsep{-1mm}
		\item Coherent
		\item Informative
		\item Non-Redundant
		\item Length was about right
	\end{itemize}
\end{itemize}
The average and (StD) results on the 12 sessions (2 topic for 6 participants) are:
\input{figures/tabUserStudyScores}

\paragraph{Comments raised by participants.}
\label{sec_appendix_user_study_issues}

During the sessions, the experiment overseer collected comments and ideas for improvements raised by the participants. The consensus was that the summaries were very impressive, especially when realizing that they summarize many sentences from multiple documents, and that the \textit{Concepts} and \textit{Entities} facets were useful for navigating through the vast information.
For improvement, suggestions included to reverse the order of the history list, to add a reset button of all filters and to move the facet-value frequency meter closer to the facet-value label. Some mentioned that the \textit{Statements} facet was less useful, since it acts like a summary that is unnecessary with respect to the navigation process.

\subsection{System Comparison Experiment}
\label{sec_appendix_comparison_experiment}
For the comparative experiment, we gathered 4 graduate students from our NLP lab and gave them offline assignments which took about 45 minutes. 
At the beginning, each student was given a document of instructions describing \App{} and DocFetcher, and were told to take a few minutes to play with each system on a third topic. Then each student was given a document with an assignment, with the same tasks as the usability study (Table \ref{table_experiments_tasks}). The participants were told to stop the exploration process once they felt satisfied with their outcome. There were 4 variants of the assignment document (one for each student), for all combinations of 2 systems and 2 topics, where a participant does not repeat the topic on both systems.

\paragraph{Questionnaire.}

After both topics, the users answered a comparative usability questionnaire, as mentioned in \sect \ref{sec_experiments_timing}.

The average time for completing the assignment with DocFetcher was 16 minutes and the average for \App{} was 15 minutes. We found that drafts written by participants using the two systems were comparable in informativeness, and importantly that the participants preferred \App{} over the standard search approach (from questionnaire results and general comments).

\subsection{Summary Quality Assessment}
\label{sec_appendix_summaryAssessment}

To assess the quality of the summaries output by our system (using BART fine-tuned on a summarization task), we collected 5 output summaries from each of the 6 supported topics (30 summaries overall) by submitting random facet-value selections (one or more selections per summary). These selections yielded sentence sets (summarizer inputs) of varying sizes (2 to 47 sentences).

The summaries were rated for five standard summary readability criteria, as defined in \citep{dang2006DUCoverview},
on a 1-to-5 Likert scale. Two of the authors rated all summaries, and then reconciliated on scores with a large (3 or more points) difference, in which case scores may have been slightly revised.
In addition, we added a sixth aspect - ``Factuality'', which was assessed by binary scoring. For each of the 30 summaries, a single sampled summary sentence was scored 0 if any fact in it did not have evidence in the source text, and 1 otherwise (30 sentences tested). We found that many sentences were lightly paraphrased or were fusions of two sub-sentential extractions, yielding high factuality scores. Results appear in Table \ref{table_summarization_quality}.

\input{figures/tabSummarizationQuality}

\section{\App{} Features and Sample Session}
\label{sec_appendix_interface_features}

\paragraph{Feature Explanations.}
Some of the features of \App{}, presented in \sect\ref{sec_interface}, are further explained here:
\begin{itemize}
\addtolength\itemsep{-2mm}
    \item Clicking ``Original sentences'' for a summary opens a pop-up window with the set of sentences used to generate the summary. The sentences are marked with mentions pertaining to the selected facet-values. They are grouped by their parent document and then listed in order of their position in their corresponding document. (Figure \ref{fig_orig_sentences})
    \item Clicking a document title in the sentences pop-up opens another pop-up window with that document in full. The sentences from the parent pop-up are marked in red. (Figure 
    \ref{fig_orig_document})
    \item Clicking ``History'' opens a pop-up window with all the facet-value selections and resulting summaries from the current exploration session. (Figure \ref{fig_history})
    \item If a complete sentence from the summary has already been seen in a previous one, that sentence is tinted in purple. We found this useful given the summarization model's occasional extractiveness. (Figure \ref{fig_purple_sentence})
\end{itemize}

\input{figures/figOrigSentences}
\input{figures/figOriginalDocument}
\input{figures/figHistory}
\input{figures/figPurpleSentence}

\paragraph{Facet-value examples.}
We show in Table \ref{table_facet_values_examples} some examples of facet-values and their mention clusters.

\input{figures/tabFacetValuesExamples}

\paragraph{Sample session.}
In Table \ref{table_userstudy_session} we show the facet selections and resulting summaries from part of a session in the usability study.

\input{figures/tabUserStudySession}

%% file: figures/tabExperimentsTasks.tex
\begin{table}[h!]
\begin{tabularx}{\columnwidth}
{
 | > {}l 
 | > {}X |
}
\hline
\textbf{Topic}      & \textbf{Task}                                                                                                                              \\ \hline
\makecell[tl]{Native American \\ Challenges\\(D0601)}  &  As a junior reporter, you were assigned a task to read 25 documents about Native American Challenge and hand out a draft to a reporter who will write the actual report.
\newline
For your draft, describe two / three challenges that Native American communities face. For each challenge, explain any possible causes, difficulties that arise, and things being done for or against.
 \\ \hline
\makecell[tl]{EgyptAir Crash\\(D0617)}   & As a junior reporter, you were assigned a task to read 25 documents about the EgyptAir Crash and hand out a draft to a reporter who will write the actual report.
\newline
Describe the crash and two theories around it. For each theory, describe who stands behind it, who opposes it and what are the claims supporting it.
\\ \hline
\end{tabularx}
\caption{The tasks that each user received in both usability study and comparison study. The tasks order was shuffled among the users. }
\label{table_experiments_tasks}
\end{table}

%% file: figures/tabSUSIndividual.tex
\begin{table}[h!]
\begin{tabularx}{\columnwidth}
{|X|c|c|c|c|c|c|}
\hline
\textbf{User}  & \textbf{1} & \textbf{2} & \textbf{3} & \textbf{4} & \textbf{5} & \textbf{6} \\ \hline
\textbf{Score} & 82.5       & 85         & 50 & 97.5 & 100 & 82.5 \\ \hline
\end{tabularx}
\end{table}

%% file: figures/tabUserStudyScores.tex
\begin{table}[h!]
\centering
\begin{tabularx}{0.9\columnwidth}
{|X|c|c|}
\hline
\textbf{System Aspect}        & \textbf{Score} \\ \hline
Facets quality                          & 4.3 (0.7)                         \\ \hline
Summ. coherence                         & 4.7 (0.5)                         \\ \hline
Summ. informativeness                   & 4.2 (1.1)                         \\ \hline
Summ. non-redundancy                    & 3.8 (1.0)                         \\ \hline
Summ. length is about right             & 4.3 (0.9)                         \\ \hline
\end{tabularx}
\end{table}

%% file: figures/tabSummarizationQuality.tex
\begin{table}[ht]
\begin{tabularx}{\columnwidth}
{|X|c|c|}
\hline
\textbf{Summary Aspect}        & \textbf{Score} \\ \hline
Grammatically                 & 4.20 (0.94)    \\ \hline
Non-redundancy                & 4.58 (0.77)    \\ \hline
Referential clarity           & 4.10 (1.02)    \\ \hline
Focus                         & 3.93 (1.12)    \\ \hline
Structure \& Coherence        & 3.55 (1.14)    \\ \hline
Factuality                    & 93.3\%           \\ \hline
\end{tabularx}
\caption{Average and (StD.) scores of the summary evaluation ratings over 30 random summaries generated by the system, with a 1 (worst) to 5 (best) scale.  For Factuality, the score is the percent of factual sentences (out of 30 sentences). }
\label{table_summarization_quality}
\end{table}

%% file: figures/figOrigSentences.tex
\begin{figure}[ht]
    \centering
    \includegraphics[width=\columnwidth]{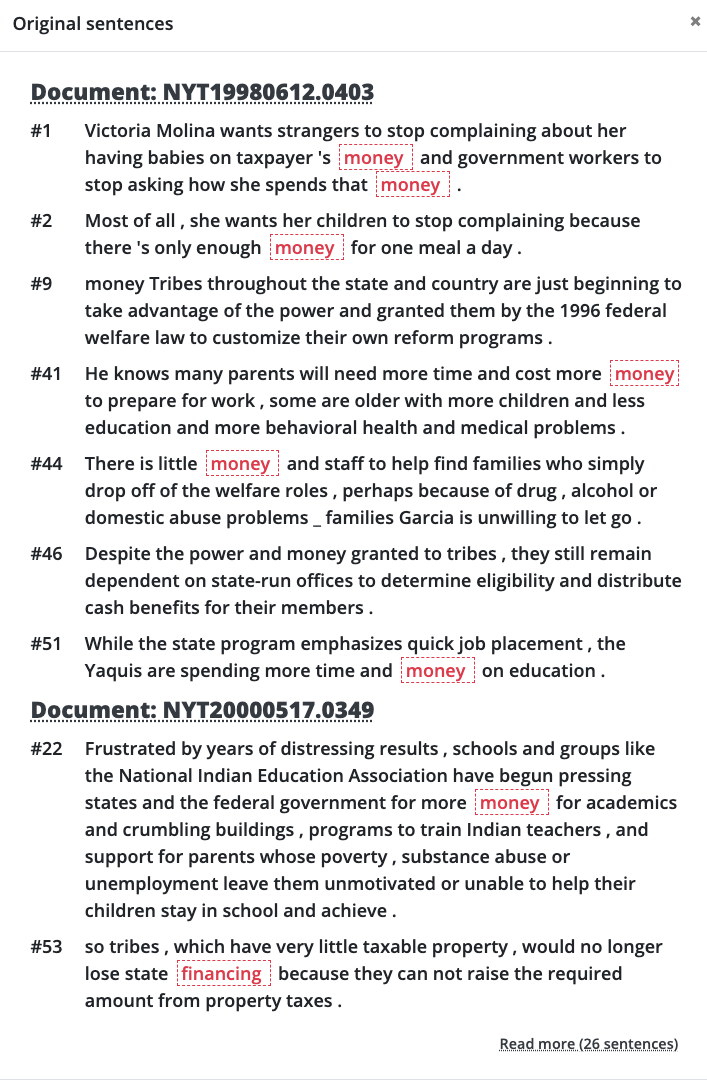}
    \caption{The original sentences popup, which lists the sentences used to create the inquired summary.}
    \label{fig_orig_sentences}
\end{figure}

%% file: figures/figOriginalDocument.tex
\begin{figure}[ht]
    \centering
    \includegraphics[width=\columnwidth]{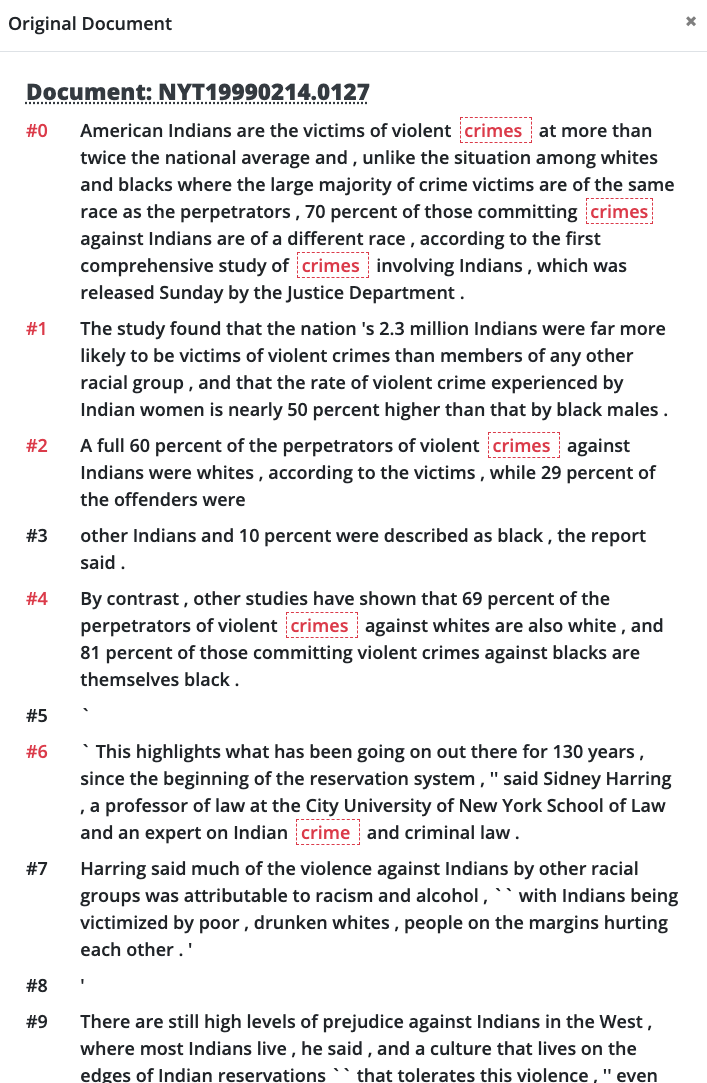}
    \caption{The original document popup, marking the sentences and mentions relevant to the \textit{sentences popup} from which this document was requested. A document enables the user to get more context on the summary.}
    \label{fig_orig_document}
\end{figure}

%% file: figures/figHistory.tex
\begin{figure}[ht]
    \includegraphics[width=\columnwidth]{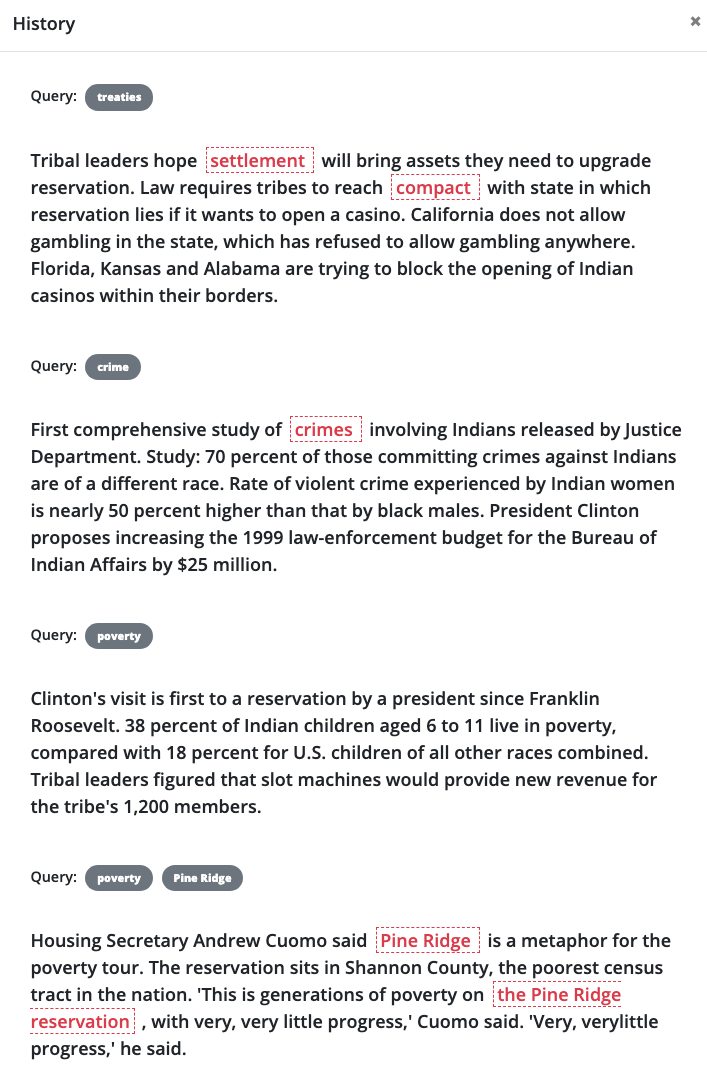}
    \caption{The history popup, containing all interactions from the current session. For each interaction, the facet selections and corresponding summary are shown.}
    \label{fig_history}
\end{figure}

%% file: figures/figPurpleSentence.tex
\begin{figure}[ht]
    \centering
    \includegraphics[width=\columnwidth]{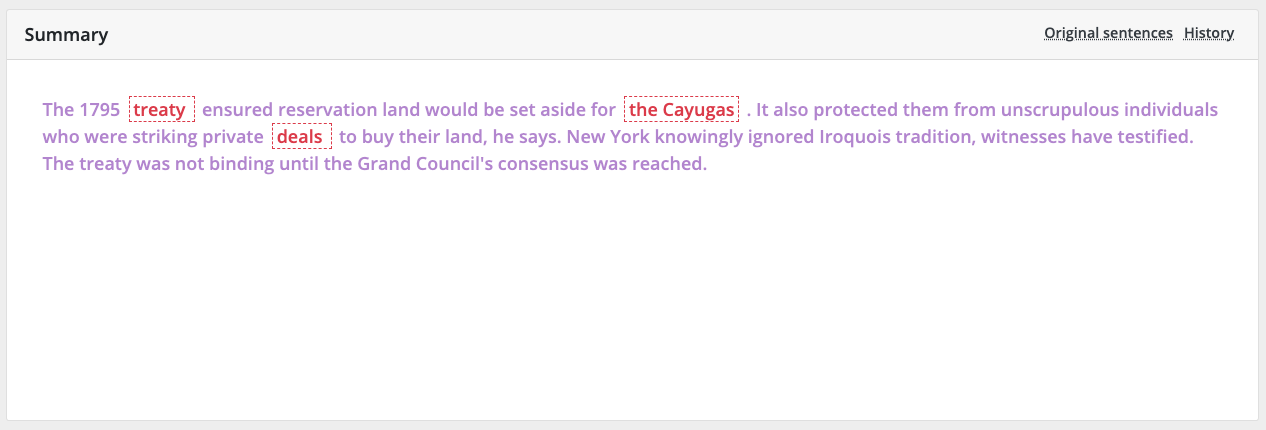}
    \caption{The sentence is tinted purple, indicating it was already extracted as part of a previous summary, relieving the user from reading it again.}
    \label{fig_purple_sentence}
\end{figure}

%% file: figures/tabFacetValuesExamples.tex
\begin{table*}[t]
\begin{tabularx}{\linewidth}
{
 | > {}l 
 | > {}X |
}
\hline
\textbf{Facet}      & \textbf{Facet-Value Examples}                                                                                                                              \\ \hline
Concepts   & ``treaties'' (``agreements'', ``deals'', ``treaty'', ``deal'', ``settlements'', ...)\newline``revenues'' (``incomes'', ``profit'') \\ \hline
Entities   & ``Makah'' (``The Makah tribe'', ``The Olympic Peninsula tribe'', ...)\newline``the plane'' (``the jet'', ``767'', ``EgyptAir Flight 990'') \\ \hline
Statements & ``Native Americans are leaving reservations and relocating in urbans areas'' (``Indians now living in urban areas'', ``migration from the reseravtions continues'', ...) \\ \hline
\end{tabularx}
\caption{Examples of facet-values. }
\label{table_facet_values_examples}
\end{table*}

%% file: figures/tabUserStudySession.tex
\begin{table*}[t]
\centering
\begin{tabularx}{\linewidth}
{
 | > {}l 
 | > {}X |
}
\hline
\textbf{Query} & \textbf{Summary}                                                                                                                              \\ \hline
\makecell[tl]{treaties\\(34 sentences)}  & Tribal leaders hope \textbf{settlement} will bring assets they need to upgrade reservation. Law requires tribes to reach \textbf{compact} with state in which reservation lies if it wants to open a casino. California does not allow gambling in the state, which has not allowed gambling in Nebraska. Florida, Kansas and Alabama have sued the U.S. Interior Department. \\ \hline
\makecell[tl]{treaties, New York\\(5 sentences)}  & McCurn previously ruled that \textbf{New York} illegally acquired the Cayugas reservation land in 1795 and 1807. The state purchased it in violation of the 1790 Indian Trade and Intercourse Act, which required Congressional approval for all Indian land \textbf{transactions}. It was long-standing New York policy to assume authority over Indian land \textbf{deals} within its borders. \\ \hline
\makecell[tl]{treaties, Florida\\(1 sentence)}  & In addition, \textbf{Florida}, Kansas and Alabama, trying to block the opening of Indian casinos within their borders, have sued the U.S. Interior Department with the aim of overturning new rules that allow the federal government to license tribal casinos in cases where states are reluctant to negotiate \textbf{compacts}. \\ \hline
\end{tabularx}
\caption{A snippet of a sample \App{} session. Words in bold are mentions of the selected facet-value(s) (e.g., "compact" is a mention of "treaties").}
\label{table_userstudy_session}
\end{table*}